\documentclass[runningheads]{llncs}

\usepackage[T1]{fontenc}
\usepackage{graphicx}
\usepackage{amsmath,amssymb}
\usepackage{multirow}
\usepackage{pifont}
\newcommand{\xmark}{\ding{55}}
\usepackage{booktabs}

\usepackage{hyperref}
\hypersetup{hidelinks}
\usepackage{orcidlink}
\renewcommand{\orcidID}[1]{\orcidlink{#1}}
\begin{document}
\title{Revitalizing Dense Material Segmentation: Stabilized Vision Transformers and the Generalization Paradox}
\titlerunning{Revitalizing Dense Material Segmentation}
\author{Allan Kazakov\inst{1,2} \orcidID{0009-0002-2122-0064} \and
Duygu Cakir\inst{1,3} \and
Hilal Kurt İrfanoğlu\inst{2} \and
Yavuz İrfanoğlu\inst{2}}

\authorrunning{A. Kazakov et al.}

\institute{Bahcesehir University, Istanbul, Turkey \\
\email{allan.kazakov@bahcesehir.edu.tr}, \email{duygu.cakir@bau.edu.tr} \and
Poder Bilişim Teknolojileri Sanayi ve Ticaret A.Ş., Istanbul, Turkey
\\
\email{hilal@poder.com.tr}, \email{yavuz@poder.com.tr} \and
Galatasaray University, Istanbul, Turkey}
\maketitle              
\begin{abstract}
Material segmentation---the pixel-wise classification of physical surface properties remains a challenging problem in computer vision, requiring physicochemical understanding distinct from object-centric parsing. Despite the introduction of the rigorous Apple Dense Material Segmentation (DMS) dataset, the benchmark has suffered from attrition and stagnation, increasingly overshadowed by geometry-biased foundation models. In this paper, we revive the Apple-DMS benchmark to establish a modern Vision Transformer baseline. We conduct an exhaustive evaluation of SegFormer and Mask2Former architectures, revealing that standard training paradigms fail on amorphous texture fields due to high-variance gradients. To address this, we introduce a stabilized training recipe featuring High-Fidelity Logit Projection, Query Entropy Regularization, and a domain-specific, physics-compliant augmentation pipeline. Our optimized SegFormer-B5 achieves a new State-of-the-Art (SOTA) of 0.4572 mIoU on the original dataset split, significantly surpassing the prior convolutional baseline. Furthermore, we identify a critical ``Generalization Paradox'': while re-partitioning the dataset into a data-rich 80/10/10 split inflates the metric to 0.5276 mIoU, expert qualitative analysis reveals this induces distributional homogenization, severely degrading real-world, out-of-distribution performance. By releasing our recovered dataset index and robust training framework, we demonstrate that material perception is far from solved and urge the community to leverage the rigorous original split to drive genuine progress in physically grounded artificial intelligence.

\end{abstract}

\section{Introduction}
\label{sec:intro}

Visual understanding is not limited to recognizing discrete objects (``things'') 
such as cars, chairs, or pedestrians; it also requires identifying the physical 
materials that compose them. Distinguishing concrete from stone, fabric from 
leather, or polished wood from laminated plastic is essential for robotic 
manipulation, photorealistic augmented reality, acoustic simulation, and scene 
understanding. Unlike standard semantic segmentation, material segmentation 
depends less on object shape and more on high-frequency texture, reflectance, 
and surface appearance cues.

The Apple Dense Material Segmentation (DMS) dataset \cite{apple_dms_2022} 
remains one of the most rigorous benchmarks for this task, providing dense 
annotations over 46 material classes. Yet the benchmark has seen surprisingly 
little progress since its introduction, with the original ResNet-50 baseline 
still serving as the main point of comparison. At the same time, the field has 
shifted toward foundation models such as SAM 2 \cite{ravi2024sam2} and SAM 3 
\cite{feichtenhofer2025sam3}, which are highly effective at separating 
geometric regions but do not directly solve material recognition. A model may 
segment a tabletop accurately while still failing to decide whether it is wood, 
plastic, glass, or polished stone.

In this work, we revisit Apple-DMS and establish a modern Vision Transformer 
baseline for dense material segmentation. We evaluate two representative 
segmentation paradigms: \textbf{Mask2Former} \cite{cheng2022masked}, which 
formulates segmentation as mask classification, and \textbf{SegFormer} 
\cite{xie2021segformer}, which uses a hierarchical semantic encoder. Our 
experiments show that applying these architectures off the shelf is 
insufficient. Dense material fields produce unstable optimization, 
high-variance gradients, and rapid overfitting, especially because material 
boundaries are often gradual rather than object-like.

To address these issues, we introduce a stabilized training recipe built around 
\textit{High-Fidelity Logit Projection}, \textit{Query Entropy Regularization}, 
differential learning rates, and a physics-aware augmentation pipeline. With 
this recipe, SegFormer-B5 achieves a new state of the art of 
\textbf{0.4572 mIoU} on the original Apple-DMS split, surpassing the prior 
convolutional baseline of \textbf{0.4200 mIoU}.

We further study whether the difficulty of Apple-DMS is caused by limited 
training data. A stratified 80/10/10 re-partition raises SegFormer-B5 
performance to \textbf{0.5276 mIoU}, but expert evaluation on real-world 
out-of-distribution samples reveals worse material behavior. We call this the 
\textit{Generalization Paradox}: a data-rich IID split can inflate benchmark 
metrics while weakening real-world material generalization. This finding 
suggests that the original Apple-DMS split remains important precisely because 
it is difficult.

Our core contributions are summarized as follows:
\begin{enumerate}
    \item We establish a new SegFormer-B5 state of the art on the original 
    Apple-DMS split, achieving \textbf{0.4572 mIoU} and outperforming the 
    previous convolutional baseline.
    
    \item We identify optimization failures that arise when training modern 
    Transformers on dense material fields and introduce a stabilization recipe 
    based on \textit{High-Fidelity Logit Projection}, 
    \textit{Query Entropy Regularization}, and differential optimization.
    
    \item We propose a \textit{Texture-First} augmentation strategy and show 
    that its effect depends strongly on the evaluation split.
    
    \item We identify the \textit{Generalization Paradox}, showing that 
    data-rich IID splits can inflate mIoU while reducing out-of-distribution 
    robustness, and release recovered dataset indices, code, and training 
    recipes to support reproducible future work: 
    \href{https://github.com/AllanK24/Material_Seg_AppleDMS}
    {\texttt{github.com/AllanK24/Material\_Seg\_AppleDMS}}.
\end{enumerate}

\section{Related Work}
\label{sec:related}

\paragraph{Material segmentation and texture analysis.}
Material perception has long been treated as distinct from object recognition: 
where object-centric segmentation identifies discrete ``things,'' material 
segmentation must classify the surface properties of ``stuff'' 
\cite{adelson2001seeing}. Early work focused on texture and material 
classification using image-level labels, leading to datasets such as the 
Describable Textures Dataset (DTD) \cite{cimpoi2014describing} and the Flickr 
Material Database (FMD) \cite{sharan2013recognizing}. These resources are 
valuable for recognizing material appearance, but they do not provide dense 
spatial supervision. Later benchmarks such as OpenSurfaces 
\cite{bell2013opensurfaces} and Materials in Context (MINC) 
\cite{bell2015material} introduced spatial material annotations, but primarily 
through point-level labels. The Apple Dense Material Segmentation (DMS) dataset 
\cite{apple_dms_2022} addressed this limitation by providing dense annotations 
over 46 material classes, making it a uniquely rigorous benchmark for full-scene 
material parsing. Unlike broader segmentation datasets such as COCO-Stuff 
\cite{caesar2018coco} and ADE20K \cite{zhou2017scene}, Apple-DMS focuses 
directly on physicochemical surface categories such as \textit{Plastic}, 
\textit{Ceramic}, \textit{Wood}, and \textit{Polished Stone}. Despite this, it 
has seen limited follow-up work since its introduction.

\paragraph{Segmentation architectures.}
Convolutional architectures such as Fully Convolutional Networks 
\cite{long2015fully} and DeepLab \cite{chen2017deeplab} established the 
standard paradigm for dense prediction, using local filters and dilated 
convolutions to preserve spatial resolution while enlarging the receptive field. 
Vision Transformers \cite{dosovitskiy2020image} introduced global self-attention, 
which is attractive for material recognition because surface identity often 
depends on non-local cues such as illumination, reflectance, and repeated 
texture. Modern Transformer-based segmentation methods largely follow two 
directions. Hierarchical semantic encoders such as SegFormer 
\cite{xie2021segformer} preserve multi-scale spatial detail, while mask-based 
models such as Mask2Former \cite{cheng2022masked} formulate segmentation as a 
set prediction problem over learned mask queries. Both approaches perform well 
on object- and scene-centric benchmarks, but their behavior on dense, 
texture-heavy material fields remains underexplored.

\paragraph{Foundation models and the material gap.}
Foundation segmentation models such as SAM \cite{kirillov2023segment} and its 
successors \cite{ravi2024sam2,feichtenhofer2025sam3} have shown impressive 
ability to separate visual regions using geometric and semantic cues. However, 
accurate region extraction is not equivalent to material understanding. Prior 
analyses have shown that SAM-style models can struggle on real-world cases where 
region boundaries are ambiguous or low-contrast \cite{ji2023sam_investigation}. 
For material parsing, the core challenge is not only to outline a table, wall, 
or floor, but to determine whether a surface is wood, laminate, glass, ceramic, 
or polished stone. This requires dense supervision over material categories, 
which motivates revisiting supervised specialists trained on Apple-DMS rather 
than relying solely on geometry-first foundation models.

\section{Methodology}
\label{sec:method}

We describe the pipeline used to evaluate modern Transformer architectures on 
Apple-DMS: dataset recovery, split construction, model configurations, 
optimization stabilizers, and material-specific augmentation.

\subsection{Dataset Recovery and Preparation}
\label{subsec:dataset}

The original Apple-DMS release referenced images hosted on Flickr and AWS, some 
of which are no longer accessible due to link rot. We recovered 41,396 of the 
intended 44,560 images, corresponding to a 92.9\% recovery rate. The remaining 
images failed due to expired or unavailable URLs, request-limit errors, network 
failures, or corrupt payloads.

To support reproducibility, we release the recovered dataset indices and split 
configurations on HuggingFace. The Original and Custom splits are available as 
\href{https://huggingface.co/datasets/AllanK24/apple-dms-materials}
{\texttt{AllanK24/apple-dms-materials}} and 
\href{https://huggingface.co/datasets/AllanK24/apple-dms-materials-v2}
{\texttt{AllanK24/apple-dms-materials-v2}}, respectively.

\subsection{Data Splitting Strategy}
We investigate two distinct data splitting strategies to analyze the impact of training volume on material generalization.

\textbf{1. The Original Split (54/23/23):} The authors of Apple-DMS proposed a highly rigorous split: 22,492 training images (54\%), with a massive 46\% of the data reserved for validation and testing. This distribution was designed to test extreme generalization but severely limits the data available for hungry Transformer models.

\textbf{2. The Stratified Custom Split (80/10/10):} We hypothesize that the poor performance of modern baselines may be attributed to data starvation. We propose a new stratified split that allocates 80\% of the data (33,118 images) to training, while maintaining class balance across the validation (10\%) and test (10\%) sets. We utilize Jensen-Shannon Divergence (JSD) to verify that the pixel-level class distributions in our custom validation and test sets align closely with the training set (JSD $\approx$ 0.015), ensuring a valid statistical evaluation.

\subsection{Network Architectures}
We benchmark two architectures that represent opposing philosophies in segmentation:

\textbf{SegFormer-B5 (Semantic Encoder):} We utilize the largest variant of the SegFormer family \cite{xie2021segformer}, initialized with weights pre-trained on ADE20K. SegFormer avoids positional encodings, using a hierarchical Mix Transformer (MiT) encoder ($B5$) that relies on overlapping patch merging. We hypothesize this inductive bias is critical for materials, as it preserves the high-frequency texture continuity (e.g., wood grain) that is often lost in standard Vision Transformers.

\textbf{Mask2Former Swin-Large (Mask Classification):} We use Mask2Former \cite{cheng2022masked} with a Swin-Large backbone, also initialized with weights pre-trained on ADE20K. This model predicts binary masks for ``stuff'' classes using object queries. While powerful for distinct objects, we analyze whether its reliance on masked attention struggles with the amorphous, boundary-less nature of material segments.

\subsection{Architectural Adaptations and Optimization Recipe}
A primary contribution of this work is the identification of a 
\textit{Stabilized Training Recipe} for material parsing. Naive use of standard 
implementations produced gradient norms above 3.7 and rapid overfitting. We 
address these instabilities through two architectural adaptations that preserve 
high-frequency texture supervision, together with differential optimization.

\paragraph{1. High-Fidelity Logit Projection (SegFormer).}
A limitation of hierarchical encoders such as SegFormer is the resolution 
bottleneck: the decoder outputs logits 
$\mathbf{Z} \in \mathbb{R}^{C \times \frac{H}{4} \times \frac{W}{4}}$, while 
ground truth masks lie in $\mathbf{Y} \in \mathbb{R}^{H \times W}$. Standard 
training aligns these by downsampling $\mathbf{Y}$ with nearest-neighbor 
interpolation, which can introduce aliasing at fuzzy material boundaries and 
discard fine texture cues such as \textit{Fabric} weave or \textit{Wood} grain.

We therefore introduce \textbf{High-Fidelity Logit Projection}. Instead of 
degrading the supervision signal, we project logits into the high-resolution 
pixel space by bilinear interpolation $\mathcal{U}(\cdot)$ before computing the 
loss, and apply a resolution-aware label smoothing objective:
\begin{equation}
    \mathcal{L}_{HFLP} = -\frac{1}{HW} \sum_{i,j} \sum_{c=1}^{C} 
    \left[ (1-\epsilon)y_{i,j,c} + \frac{\epsilon}{C} \right] 
    \log \left( \sigma(\mathcal{U}(\mathbf{Z}))_{i,j,c} \right)
\end{equation}
where $\sigma$ is the softmax function and $\epsilon=0.1$ balances robustness to 
fuzzy transitions with preservation of semantic identity.

\paragraph{2. Query Entropy Regularization (Mask2Former).}
For Mask2Former, applying standard label smoothing directly conflicts with the 
Hungarian matching objective, whose cost matrix depends on exact class 
probabilities. Modifying this cost with soft labels destabilizes query-to-mask 
assignment. To avoid this, we introduce \textbf{Query Entropy Regularization 
(QER)}, an auxiliary term applied to the class query logits $\mathbf{Q}$ after 
matching. We model the smoothing prior as a uniform distribution $U$ and 
regularize the predicted query distribution by:
\begin{equation}
    \mathcal{L}_{total} = \mathcal{L}_{match} + \lambda \cdot 
    D_{KL}\Big( \log(\sigma(\mathbf{Q})) \parallel U \Big)
\end{equation}
where $\lambda=0.1$. This maximum-entropy constraint discourages early 
over-confident collapse while preserving the bipartite matching kernel. In our 
experiments, removing QER caused rare material classes to be ignored in favor of 
dominant background materials, while larger values over-penalized certainty and 
reduced mask precision.

\paragraph{3. Differential Learning Rates and Scheduling.}
To prevent catastrophic forgetting of ImageNet/ADE20K pretraining, we optimize 
backbone and decoder parameters with separate learning rates: 
$\eta_{back}=1 \times 10^{-4}$ and 
$\eta_{head}=1 \times 10^{-3}$. We also replace linear decay with cosine 
annealing to a hard minimum of $\eta_{min}=1 \times 10^{-6}$, which keeps late 
training updates active enough to refine fine-grained texture boundaries.

\subsection{Domain-Specific Augmentation Pipeline}
\label{subsec:augmentation}

Object-centric augmentations can distort material identity, particularly when 
they alter color, reflectance, or fine texture. We therefore use a 
\textit{Texture-First} augmentation pipeline with three constraints. First, 
Large Scale Jittering encourages invariance to scene context while preserving 
local texture structure. Second, photometric perturbations are deliberately 
bounded, especially hue, to avoid changing perceived material identity. Third, 
we add synthetic specular highlights to improve robustness to view-dependent 
reflection, and apply Gaussian ISO noise only to SegFormer, since high-frequency 
noise destabilized Mask2Former's query-based masks in our experiments. The full 
hyperparameter configuration is given in Table~\ref{tab:aug_params}.

\begin{table}[h]
\caption{Domain-Specific Augmentation Hyperparameters.}\label{tab:aug_params}
\centering
\begin{tabular}{llcc}
\toprule
\textbf{Category} & \textbf{Parameter} & \textbf{Mask2Former} & \textbf{SegFormer} \\
\midrule
\multirow{3}{*}{Geometric} & Scale Range (LSJ) & $[0.1, 2.0]$ & $[0.5, 2.0]$ \\
 & Crop Size & $512 \times 512$ & $512 \times 512$ \\
 & Horizontal Flip & $p=0.5$ & $p=0.5$ \\
\midrule
\multirow{5}{*}{Photometric} & Hue Delta & $0.02$ & $0.02$ \\
 & Saturation Range & $[0.8, 1.2]$ & $[0.8, 1.2]$ \\
 & Contrast Range & $[0.6, 1.4]$ & $[0.7, 1.3]$ \\
 & Specular Injection & $p=0.3$ & $p=0.3$ \\
 & Gaussian Noise & \textbf{Disabled} & $p=0.3$ \\
\bottomrule
\end{tabular}
\end{table}

\section{Experiments and Results}
\label{sec:experiments}

We evaluate SegFormer-B5 and Mask2Former Swin-L on the recovered Apple-DMS 
benchmark under two data partitioning strategies: the Original Split and our 
Stratified Custom Split. All models use AdamW optimization, differential 
learning rates for backbone and decoder parameters, and cosine annealing with a 
hard minimum learning rate. Unless otherwise stated, performance is reported 
using mean Intersection-over-Union (mIoU), mean accuracy (mAcc), and overall 
accuracy (aAcc).

\subsection{Implementation Details}

Table~\ref{tab:hyperparams} summarizes the shared optimization configuration. 
For the Original Split, convergence plateaued after 20 epochs. For the larger 
Custom Split, we trained for 40 epochs to account for the increased number of 
training images.

\begin{table}[h]
\caption{Unified hyperparameter configuration. $\epsilon$ denotes the 
SegFormer smoothing coefficient used in High-Fidelity Logit Projection, while 
$\lambda$ denotes the Mask2Former Query Entropy Regularization coefficient.}\label{tab:hyperparams}
\centering
\begin{tabular}{l|c}
\toprule
\textbf{Hyperparameter} & \textbf{Value (Unified)} \\
\midrule
Optimizer & AdamW \\
Batch Size & $256$ \\
Weight Decay & $0.1$ \\
Regularization coefficient & $\epsilon=0.1$ / $\lambda=0.1$ \\
\midrule
\textbf{Learning Rates} & \\
\hspace{3mm} Backbone ($LR_{back}$) & $1.0 \times 10^{-4}$ \\
\hspace{3mm} Decoder/Head ($LR_{head}$) & $1.0 \times 10^{-3}$ \\
\hspace{3mm} Min LR ($\eta_{min}$) & $1.0 \times 10^{-6}$ \\
\midrule
\textbf{Epochs} & \\
\hspace{3mm} Original Split & 20 \\
\hspace{3mm} Custom Split & 40 \\
\bottomrule
\end{tabular}
\end{table}

\subsection{Results on the Original Apple-DMS Split}

Table~\ref{tab:main_results} reports performance on the Original Split. 
SegFormer-B5 without augmentation achieves the best result, reaching 
\textbf{0.4572 mIoU} and surpassing the original ResNet-50 Apple-DMS baseline 
of \textbf{0.4200 mIoU}. Mask2Former improves over the baseline as well, but 
falls behind SegFormer despite using a Swin-Large backbone.

Interestingly, the Texture-First augmentation pipeline does not improve results 
on the Original Split. SegFormer decreases slightly from 0.4572 to 0.4544 mIoU, 
while Mask2Former drops from 0.4410 to 0.4072 mIoU. This suggests that the 
Original Split already acts as a challenging distribution-shift evaluation, and 
that additional augmentation can move the training distribution away from the 
held-out evaluation set.

\begin{table}[t]
\caption{Test set results on the \textbf{Original Apple Split}. Aug. denotes the 
augmentation pipeline. SegFormer-B5 without 
augmentation establishes the new SOTA by mIoU.}
\label{tab:main_results}
\centering
\small
\setlength{\tabcolsep}{4pt}
\renewcommand{\arraystretch}{0.95}
\begin{tabular}{@{}lccccc@{}}
\toprule
\textbf{Method} & \textbf{Aug.} & \textbf{mIoU} & \textbf{mAcc} & \textbf{aAcc} & \textbf{mB-IoU} \\
\midrule
Apple Baseline & -- & 0.4200 & 0.5850 & 0.7290 & -- \\
\midrule
Mask2Former & \xmark & 0.4410 & 0.4912 & 0.7582 & \textbf{0.1623} \\
Mask2Former & \checkmark & 0.4072 & 0.4516 & 0.7652 & 0.1476 \\
\midrule
SegFormer & \checkmark & 0.4544 & 0.5415 & 0.8228 & 0.1342 \\
\textbf{SegFormer} & \textbf{\xmark} & \textbf{0.4572} & \textbf{0.5467} & \textbf{0.8229} & 0.1423 \\
\bottomrule
\end{tabular}
\end{table}

\subsection{Impact of the Stratified Custom Split}

Table~\ref{tab:split_comparison} shows the effect of increasing training data 
through the Stratified Custom Split. SegFormer-B5 improves substantially, 
reaching \textbf{0.5276 mIoU} with augmentation. This indicates that modern Transformer models can exploit additional dense material supervision under an IID-style evaluation split.

The effect of augmentation reverses under this split. Whereas augmentation 
reduced performance on the Original Split, it improves SegFormer from 0.5117 to 
0.5276 mIoU on the Custom Split. This indicates that augmentation behaves as a 
regularizer when the train and test distributions are closely aligned. However, 
as discussed in Sec.~\ref{sec:discussion}, the resulting metric gain does not 
necessarily translate to stronger out-of-distribution material recognition.

Mask2Former plateaued around 0.49 validation mIoU under both augmented and 
non-augmented settings, so we did not allocate final test evaluation compute to 
this model on the Custom Split.

\begin{table}[h]
\caption{Performance on the \textbf{Custom Stratified Split (80/10/10)}. Augmentation improves performance here, contrasting with the Original Split.}\label{tab:split_comparison}
\centering
\begin{tabular}{l|c|cc}
\toprule
\textbf{Method} & \textbf{Augmentation} & \textbf{Val mIoU} & \textbf{Test mIoU} \\
\midrule
Mask2Former & \xmark & 0.4944 & \textit{Not Evaluated} \\
Mask2Former & \checkmark & 0.4945 & \textit{Not Evaluated} \\
\midrule
SegFormer & \xmark & 0.5006 & 0.5117 \\
\textbf{SegFormer} & \textbf{\checkmark} & \textbf{0.5168} & \textbf{0.5276} \\
\bottomrule
\end{tabular}
\end{table}

\section{Discussion}
\label{sec:discussion}

Our experiments reveal two main findings. First, hierarchical semantic encoders 
are better aligned with dense material parsing than query-based mask classifiers. 
Second, higher IID test performance does not necessarily imply stronger 
real-world material generalization.

\subsection{Architectural Inductive Biases: Texture vs. Object Queries}

Although the gap between Mask2Former (0.4410 mIoU) and SegFormer 
(0.4572 mIoU) is modest, it reflects an important architectural mismatch. 
Mask2Former represents a scene as a set of learned mask queries, an assumption 
well suited to discrete objects but less natural for materials. A single material 
may appear in multiple disconnected regions, while adjacent materials often 
transition gradually rather than through sharp object boundaries.

SegFormer avoids explicit instance-style mask assignment. Its hierarchical MiT 
encoder progressively merges overlapping patches, preserving texture cues such 
as grain, weave, and roughness across the feature map. Our results suggest that 
material segmentation rewards semantic consistency over geometric isolation, 
which helps explain why SegFormer provides the stronger baseline on Apple-DMS.

\subsection{The Generalization Paradox: Metrics vs. Reality}

A central finding of this work is the \textit{Generalization Paradox}. We compare 
two SegFormer-B5 variants: \textbf{Model V1}, trained on the rigorous Original 
Apple Split without augmentation, which achieves \textbf{0.4572} test mIoU; and 
\textbf{Model V2}, trained on the data-rich Stratified Custom Split with 
Texture-First augmentation, which achieves \textbf{0.5276} test mIoU.

Despite its higher benchmark score, Model V2 was consistently judged worse by 
domain experts on out-of-distribution real-world samples. We hypothesize that 
the Stratified Custom Split introduces distributional homogenization: random 
80/10/10 re-partitioning makes the test set more IID with respect to training, 
allowing the model to exploit dataset-level statistical priors such as lighting, 
scene type, and photographic style. In contrast, the Original Split appears to 
function as a stricter distribution-shift evaluation, encouraging more robust 
material cues rather than split-specific correlations.

\subsection{Qualitative Analysis: Real-World Evaluation}

\begin{figure}[p] 
    \centering
    \includegraphics[width=\textwidth]{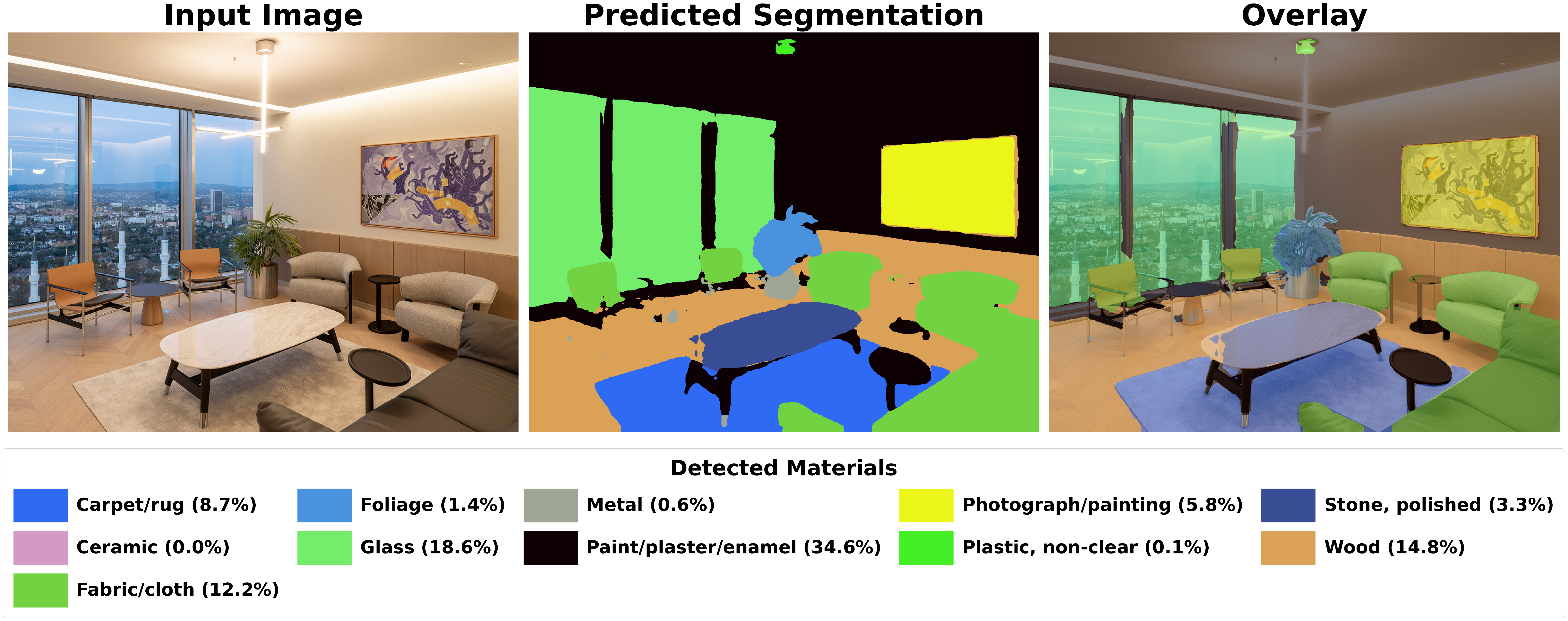} \\
    \makebox[\textwidth]{\small (a) Scene 1: \textbf{Model V1} (Original Split, 0.45 mIoU)} \\
    \vspace{2mm}
    \includegraphics[width=\textwidth]{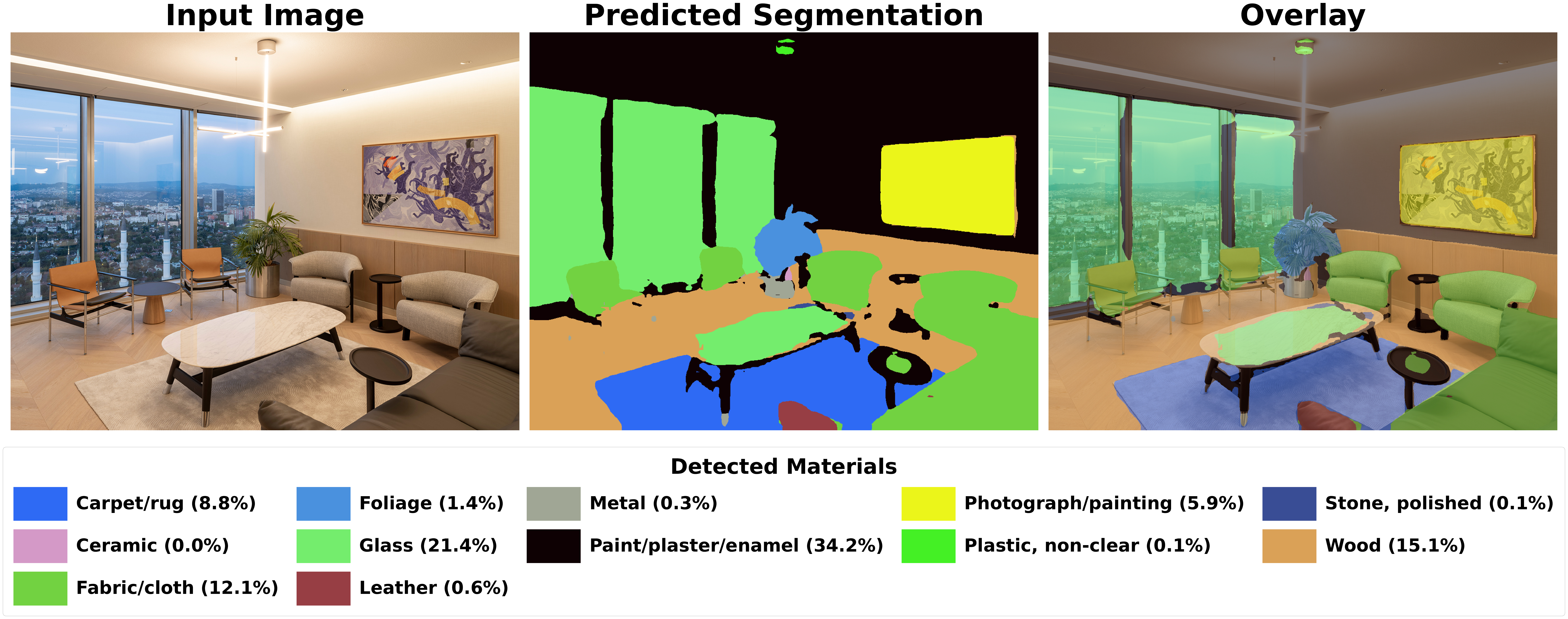} \\
    \makebox[\textwidth]{\small (b) Scene 1: \textbf{Model V2} (Custom Split, 0.52 mIoU)} \\
    \vspace{4mm}
    
    \includegraphics[width=\textwidth]{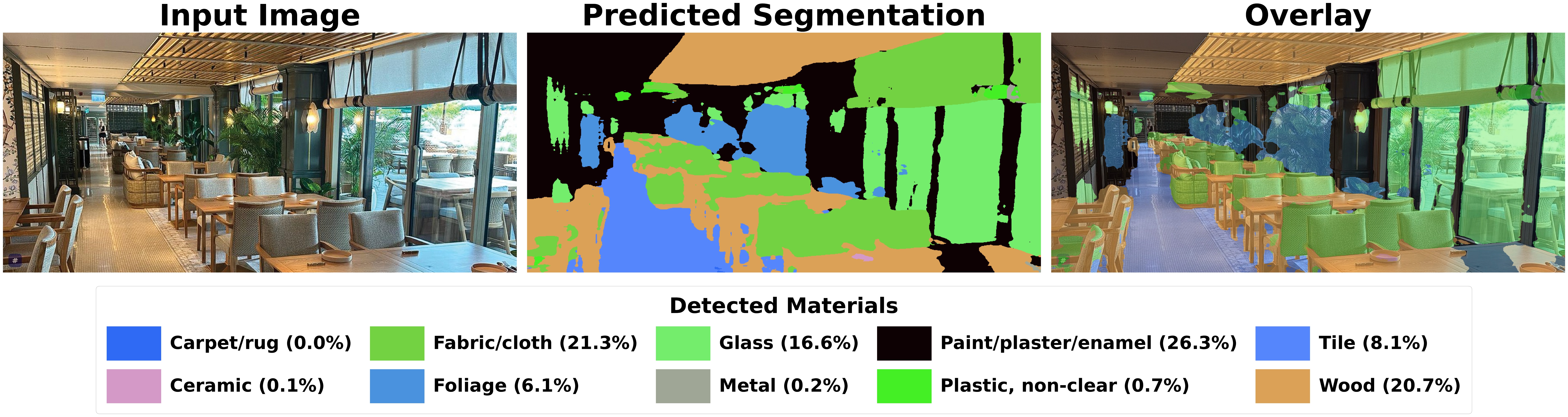} \\
    \makebox[\textwidth]{\small (c) Scene 2: \textbf{Model V1} (Original Split, 0.45 mIoU)} \\
    \vspace{2mm}
    \includegraphics[width=\textwidth]{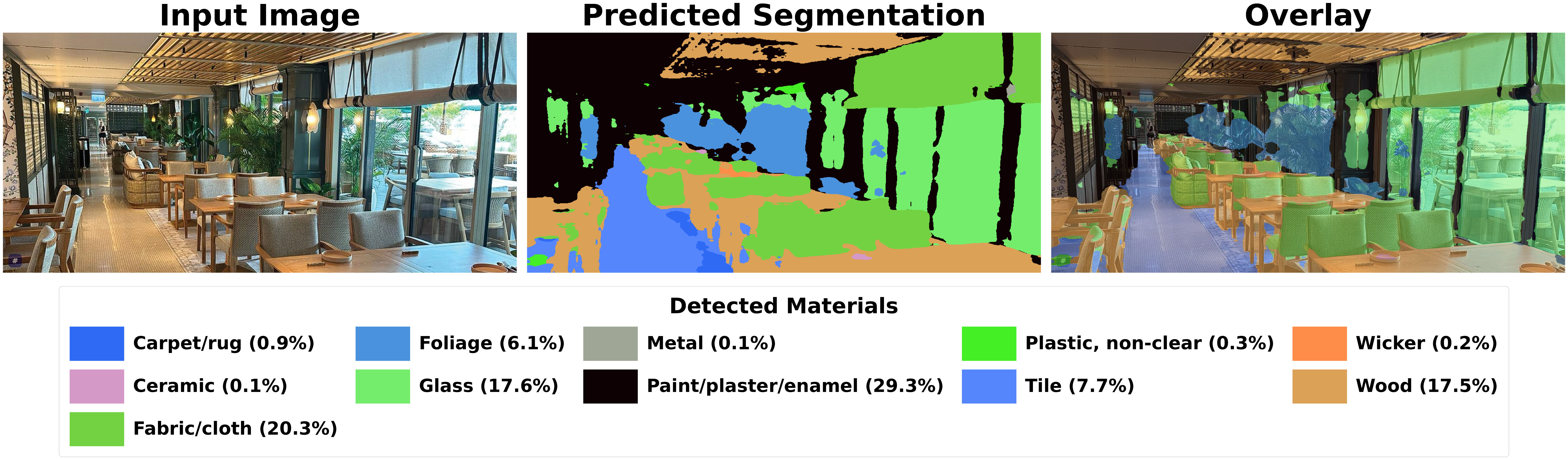} \\
    \makebox[\textwidth]{\small (d) Scene 2: \textbf{Model V2} (Custom Split, 0.52 mIoU)} \\
    
    \caption{\textbf{The Generalization Paradox on Real-World Samples.} 
Comparison of SegFormer trained on the rigorous Original Split (V1) and the 
data-rich Custom Split (V2). Although V2 achieves higher IID test mIoU, V1 
better preserves material identity under out-of-distribution conditions: 
polished stone remains robust to glare in Scene 1, and high-frequency wood and 
tile boundaries are better preserved in Scene 2.}
    \label{fig:qualitative}
\end{figure}

Figure~\ref{fig:qualitative} illustrates this discrepancy. In Scene 1, Model V1 
preserves the polished stone identity of the central table despite strong window 
glare, whereas Model V2 misclassifies the same reflective surface as 
\textit{Glass}. In Scene 2, Model V1 better preserves high-frequency wooden 
ceiling structure and cleaner tile boundaries, while Model V2 exhibits stronger 
spatial smoothing but also semantic drift, including wood being absorbed into 
\textit{Paint/plaster/enamel} and tile regions bleeding into 
\textit{Carpet/rug}. These examples show that the higher IID score of Model V2 
does not translate into stronger material understanding under real-world 
conditions.

\section{Conclusion and Limitations}
\label{sec:conclusion}
\label{sec:limitations}

This work revisits dense material segmentation through the rigorous Apple-DMS
benchmark and establishes a new state of the art of \textbf{0.4572 mIoU} using
SegFormer-B5 on the original split. Our results show that hierarchical semantic
encoders are better aligned with the continuous, texture-driven nature of
material regions than query-based mask classifiers such as Mask2Former, which
are more naturally suited to discrete object instances.

We further show that modern Vision Transformers require domain-specific
stabilization to succeed on dense material fields. High-Fidelity Logit
Projection, Query Entropy Regularization, differential learning rates, and
material-aware augmentation help mitigate the high-variance gradients,
catastrophic forgetting, and query collapse observed under naive training.

A key finding is the \textit{Generalization Paradox}: although a data-rich
80/10/10 split raises SegFormer-B5 performance to \textbf{0.5276 mIoU}, expert
evaluation on real-world samples suggests weaker out-of-distribution behavior.
This indicates that IID split construction can inflate benchmark metrics by
encouraging reliance on dataset-specific statistical priors rather than robust
material cues. The original Apple-DMS split therefore remains valuable precisely
because it is difficult.

Several limitations remain. Due to link rot, approximately 7\% of the original
Apple-DMS images could not be recovered, so comparison with the 2022 baseline is
based on the recovered subset. In addition, our out-of-distribution evaluation
relies on expert qualitative review rather than a dense ground-truth
``in-the-wild'' benchmark. Building such a benchmark is an important direction
for future work. By releasing our recovered indices, code, and training recipes,
we aim to support renewed progress on physically grounded material perception.

\begin{credits}
\subsubsection{\ackname}
This research was conducted in collaboration with Poder and supported by the
Scientific and Technological Research Council of T\"urkiye (T\"UB\.ITAK) under
the 1812 Entrepreneurship Support Program (BiGG), project no. 2240407. We thank
the domain experts at Poder for providing computational resources and
out-of-distribution qualitative evaluations.

\subsubsection{\discintname}
Allan Kazakov conducted this research during his tenure at Poder; Hilal Kurt
\.Irfano\u{g}lu is the founder of Poder; and Yavuz \.Irfano\u{g}lu is affiliated
with Poder. Duygu Cakir served as academic advisor. The authors declare no
commercial or financial relationships that could be construed as a conflict of
interest regarding the scientific findings presented in this article.
\end{credits}
%
%
%
%

\end{document}